\documentclass[5pt]{article}

\usepackage[letterpaper]{geometry}
\usepackage{spconf,amsmath,epsfig}
\usepackage{enumitem}
\usepackage[numbers,sort&compress]{natbib}

\usepackage{fancyhdr}
\begin{document}\sloppy

% Example definitions.
% --------------------
\def\x{{\mathbf x}}
\def\L{{\cal L}}

% Title.
% ------
\title{DEEP NETWORKS FOR COMPRESSED IMAGE SENSING}
%
% Single address.
% ---------------
\name{Wuzhen Shi\sthanks{This work has been supported in part by the Major State Basic Research Development Program of China (973 Program 2015CB351804), the National Science Foundation of China under Grant No. 61572155.}, Feng Jiang, Shengping Zhang and Debin Zhao}
\address{Harbin Institute of Technology, NO. 92, Xidazhi Street, Harbin, Heilongjiang  150001}
%
% For example:
% ------------
%\address{School\\
%	Department\\
%	Address\\
%   Email}
%
% Two addresses (uncomment and modify for two-address case).
% ----------------------------------------------------------
%\twoauthors
%  {A. Author-one, B. Author-two\sthanks{Thanks to XYZ agency for funding.}}
%	{School A-B\\
%	Department A-B\\
%	Address A-B}
%  {C. Author-three, D. Author-four\sthanks{The fourth author performed the work
%	while at ...}}
%	{School C-D\\
%	Department C-D\\
%	Address C-D\\
%   Email}
%

\maketitle

\begin{abstract}
The compressed sensing (CS) theory has been successfully applied to image compression in the past few years as most image signals are sparse in a certain domain. Several CS reconstruction models have been recently proposed and obtained superior performance. However, there still exist two important challenges within the CS theory. The first one is how to design a sampling mechanism to achieve an optimal sampling efficiency, and the second one is how to perform the reconstruction to get the highest quality to achieve an optimal signal recovery. In this paper, we try to deal with these two problems with a deep network. First of all, we train a sampling matrix via the network training instead of using a traditional manually designed one, which is much appropriate for our deep network based reconstruct process. Then, we propose a deep network to recover the image, which imitates traditional compressed sensing reconstruction processes. Experimental results demonstrate that our deep networks based CS reconstruction method offers a very significant quality improvement compared against state-of-the-art ones.
\end{abstract}
\section{Introduction}
\label{sec:intro}

\begin{figure}[t]
\centering
\includegraphics[width=0.4\textwidth]{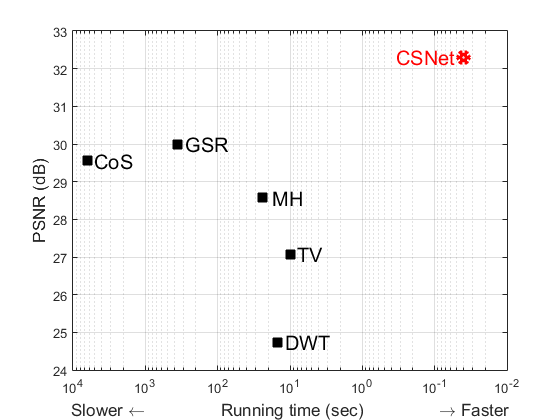}
\vspace{-10pt}
\caption{\small {The proposed CSNet achieves the state-of-the-art reconstruction quality, whilst maintains high and competitive speed in comparison to existing CS methods. The chart is based on Set5~\cite{rf23} results of 0.1 sampling ratio summarized in Table \ref{tab:tab1}.}}
\vspace{-15pt}
\label{fig:fig1}
\end{figure}

In this paper, we focus on how to sample an image signal to get a compressed one and how to efficiently recover the original image from the compressed one. The compressed sensing (CS) theory shows that if a signal is sparse or compressible, it can be accurately recovered from measurements less than that of Nyquist sampling theorem. The CS measurements are obtained through the following linear transformation
\begin{eqnarray}
y = \Phi x
\end{eqnarray}
where $y$ is an $n \times 1$  measurement vector, $x$ is the original signal with size of $N \times 1$  and $\Phi $  is an $n \times N$  sampling matrix. If $n \ll N$ , reconstructing $x$ from $y$ is generally ill-posed. In the study of CS, there are two most challenging issues including (a) the design of the sampling operator $\Phi$; (b) the development of a fast nonlinear reconstruction algorithms~\cite{rf1}. In recent years, both of them have been extensively studied.

In most works, the sampling matrix is a random matrix, for example, a Gaussian or Bernoulli matrix, which meets the Restricted Isometry Property (RIP) with a large probability. The signal can be efficiently recovered from fewer measurements sampled by the random measurement matrix. However, they always suffer some problems such as high computation cost, vast storage and uncertain reconstruction qualities. A definite matrix, such as a Toeplitz matrix~\cite{rf2} or a polynomial matrix~\cite{rf3} are other common CS measurement matrices, which need low computation cost and are easier to implement. However, their reconstruction qualities are worse than that with a random matrix. Some works design a sampling matrix for specific signals that lead to a better reconstruction result than the random matrix. For the Block Compressed Sensing (BCS), Dinh et al.~\cite{rf4} propose a structural sampling matrix to balance the conflict between the compressed ratio and reconstructed quality. In~\cite{rf5}, Gao et al. design a local structural sampling matrix by utilizing the local smooth property of images. Although a lot of works have done as discussed above, designing an effective sampling matrix is still difficult. In this paper, we design a deep network to learn a sampling matrix automatically.

Another important issue of CS is developing fast and effective nonlinear reconstruction algorithms. One kind of the CS reconstruction algorithms is convex optimization methods, which translate the nonconvex problem into a convex one to get the approximate solution. Basis Pursuit (BP)~\cite{rf6} is the most commonly used convex optimization method for compressed sampling reconstruction. It replaces the L0 norm constraint with the L1 norm one to get the solution by solving a linear programming problem. For 2D images, another well-known reconstruction algorithm is through the minimization of total variation (TV)~\cite{rf7}. To reduce the computation complexity, some fast greedy algorithms have also been proposed, such as the orthogonal matching pursuit~\cite{rf8} and the stage-wise orthogonal matching pursuit method~\cite{rf9}. As an alternative to the pursuit class of CS reconstruction, techniques based on projections have been proposed recently. In~\cite{rf1}, Lu Gan propose and study block compressed sensing for natural images, where image acquisition is conducted in a block-by-block manner through the same operator. In recent years, some other high quality compressed reconstruction methods have also been proposed. In~\cite{rf10}, Mun et al. propose a multiple hypothesis version of block compressed sensing smooth projected Landweber algorithm~\cite{rf1} with reconstruction driven by the measurement-domain residual resulting from multiple predictions culled from neighboring blocks. Zhang et al.~\cite{rf11} propose group sparse representation (GSR) to get the higher sparseness than the original signal that results in very good reconstruction performance. However, most existing work focus on the quality of the reconstruction image but ignore the computation complexity that limits their real time applications. As show in Fig. 1, these popular compressed reconstruction methods cost more than 10 seconds to several hours per image to get the high quality. In this paper, we try to propose a real time compressed reconstruction method while keeping the good performance.

Recently, deep learning method has got much attention and it is successfully applied in many high level computer vision problems. Some deep learning based methods have also been explored for the low level tasks. Dong et al.~\cite{rf12} demonstrate that a convolutional neural network (CNN) can learn a mapping from low resolution image to high resolution one in an end-to-end manner. Soon after, they expand this work for JPEG compressive image restoration~\cite{rf13}. An effective method~\cite{rf14} to reduce the amount of weights and speed it up has been proposed. Different from~\cite{rf12,rf13,rf14} that use the undegraded image as ground true for training, some works try to learn image residual. Kim et al.~\cite{rf15} propose a very deep network to learn residual to fast the convergence speed. In~\cite{rf16}, Wang et al. show that a sparse coding model particularly designed for super-resolution can be incarnated as a neural network trained in a cascaded structure from end to end. The interpretation of the network based on sparse coding leads to much more efficient and effective training, as well as a reduced model size. Motivated by the sparsity-based dual-domain method, Wang et al.~\cite{rf17} design a deep network to imitate the sparse coding process. All these previous works demonstrate deep learning is an effective method for low level computer vision problems.

In this paper, we propose a deep network to solve the two most important issues in compressed sensing, i.e. designing a sampling matrix and developing a fast nonlinear reconstruction algorithm. The traditional block compressed sensing smooth projected Landweber algorithm includes the processes of compressed sampling, initial reconstruction and non-linear signal reconstruction as shown in the upper of Fig.2, which inspires us to design a deep network with different sub-networks implementing the corresponding processes, respectively. Firstly, we use a convolution layer to imitate the process of compressed sampling, which can learn the sampling matrix automatically while avoiding complicated artificial designs. Secondly, a convolution layer of size of $1 \times 1$ and a specific combination layer, which contains the operation of reshape and concatenation, are used to implement the initial reconstruction. Finally, five convolution layers form a deep reconstruction sub-network to further improve the quality of the initial reconstructed image, which achieves the function of non-linear signal reconstruction. Experimental results indicate that the proposed method is more effective and efficient than several state-of-the-art methods as illustrated in Fig. 1.

In short, the contributions of this work are mainly in three aspects:
\begin{itemize}
  \item We establish a relationship between our deep learning based compressed sampling reconstruction and the traditional block compressed sensing smooth projected Landweber algorithm. This relationship gives insight into the design of our network structure.
  \item We design the sampling operator via a convolution layer in the deep network that avoiding complicated artificial designs.
  \item We present a convolutional neural network for compressed sampling reconstruction. The network directly learns an end-to-end mapping between the compressed measurement and the target image, and achieves good reconstruction quality and fast speed.
\end{itemize}

\section{Related Work}

\begin{figure*}[t]
\centering
\includegraphics[width=0.9\textwidth]{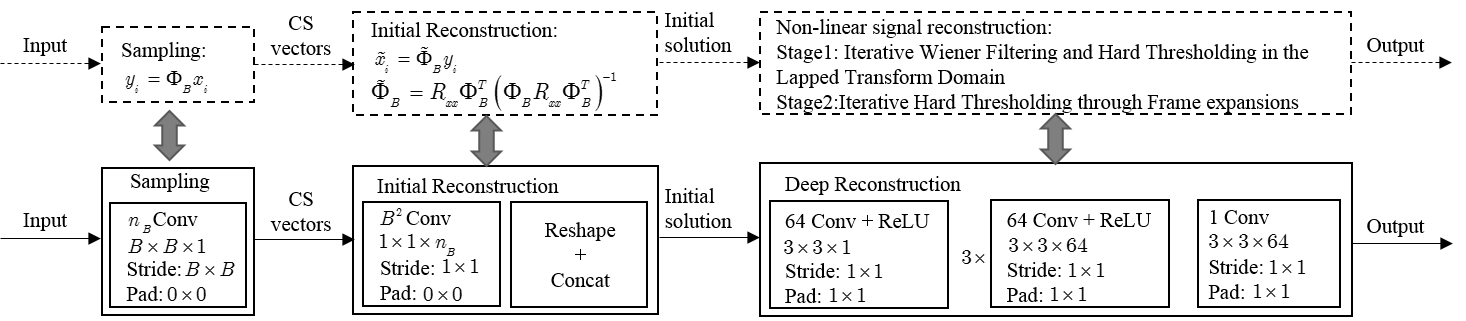}
\vspace{-12pt}
\caption{\footnotesize {The proposed compressed sensing based network structure. This figure shows comparison between our proposed CSNet and traditional BCS methods. Bottom is the framework of CSNet, while the upper is the framework of block compressed sensing of natural images proposed in~\cite{rf1}. The first layer implements block compressed sensing sampling, which learns the compressed sampling matrix automatically while avoiding complex artificial design. The second part uses a convolution layer and a combination (reshape + concat) layer to imitate the initial reconstruction process which is the minimum mean square error linear estimation in traditional BCS reconstruction. The third part is a five layer convolution network that implements the non-linear signal reconstruction process.}}
\vspace{-17pt}
\label{fig:fig2}
\end{figure*}

As an alternative to the pursuit class of CS reconstruction, techniques based on projections have been proposed recently~\cite{rf1,rf10,rf18}. This kind of algorithms obtains the reconstruction output by successively projecting and thresholding. In~\cite{rf18}, the initial solution is the result of L2 optimization, i.e. ${\tilde x^0} = {\Phi ^\dag }y$, where ${\Phi ^\dag }$ is the pseudo inverse of $\Phi $ . Then, the approximation at iteration \emph{i}+1 can be calculated as
\begin{eqnarray}
{\hat x^i} = {\tilde x^i} + \frac{1}{\gamma }\Psi {\Phi ^T}\left( {y - \Phi {\Psi ^{ - 1}}{{\tilde x}^i}} \right)
\end{eqnarray}
\begin{eqnarray}
{\tilde x^{i + 1}} = \left\{ {\begin{array}{*{20}{c}}
{{{\hat x}^i},\quad \left| {{{\hat x}^i}} \right| \ge {\tau ^i}}\\
{0\;\quad else\;\;\;\;}
\end{array}} \right.
\end{eqnarray}
where $\Psi$ is the sparsity transform domain, $\gamma$ is a scaling factor and ${\tau ^i}$ is a threshold set appropriately at each iteration.

In~\cite{rf1}, Gan proposed block compressed sensing (BCS) for natural images, which combines block based compressed sampling and smoothed projected Landweber reconstruction. In BCS, an image is divided into $B \times B$ blocks and sampled using an appropriately-sized measurement matrix. If the sampling ratio is $\frac{M}{N}$, the measurement of each block is ${n_B} = \left\lfloor {\frac{M}{N}{B^2}} \right\rfloor $ . Then ${\Phi _B}$ is a ${n_B} \times {B^2}$ orthonormal measurement matrix. Suppose ${x_j}$ is a vector representing the ${j^{th}}$ block, the corresponding measurement can be obtained as ${y_j} = {\Phi _B}{x_j}$. Different from~\cite{rf18}, Lu Gan propose to use minimum mean square error (MMSE) linear estimation to obtain the initial solution for BCS. To further improve the quality of the reconstructed images, Lu Gan propose a 2-stages non-linear reconstruction algorithm by exploiting the sparsity property. The framework of BCS proposed by Lu Gan is showed in the upper of Fig.2 for comparison with our deep learning based CS reconstruction method.

Many improved smoothed projected Landweber based BCS methods (BCS-SPL) have been proposed in the literature. For example, Mun et al. proposed a series of this kind method: MC-BCS-SPL~\cite{rf19}, MS-BCS-SPL~\cite{rf20} and MH-BCS-SPL~\cite{rf10}, which are well known in the literature since they release all the codes.

\section{Proposed Deep Network for Compressed Sensing Reconstruction}

As discussed in the above that traditional BCS-SPL methods consist of three steps including compressed sampling, initial reconstruction and non-linear signal reconstruction. Our proposed network contains the corresponding part that forms a compressed sampling sub-network and a reconstruction sub-network, which consists of an initial reconstruction sub-network and a deep reconstruction sub-network. The configuration of the proposed network is outlined in Fig. 2.

\subsection{Proposed Network}
\textbf{Compressed Sampling Sub-network}. In traditional BCS, the process of compressed sampling is expressed as ${y_j} = {\Phi _B}{x_j}$. If each row of the measurement matrix ${\Phi _B}$ is considered as a filter, we can use a convolution layer to mimic this compressed sampling process. Since the image is divided into $B \times B$ blocks, the size of each filter in the sampling layer is also $B \times B$, so that each filter outputs one measurement. For a sampling ratio $\frac{M}{N}$ , there are ${n_B} = \left\lfloor {\frac{M}{N}{B^2}} \right\rfloor $  rows in the measurement matrix ${\Phi _B}$  to obtain ${n_B}$ sampling points. Therefore, there are ${n_B}$ filters of size $B \times B \times 1$ in the sampling layer. It should be noted that the stride of the convolution layer is $B \times B$ for non-overlapping sampling as traditional BCS methods do. Furthermore, there is no biases in each filter that all the filters form a traditional measurement matrix, which can be learned automatically in the network while avoiding complicated artificial design. As in most BCS methods~\cite{rf1,rf18,rf19,rf20}, we set $B = 32$  in our experiments. Therefore, there are 102 filters in this layer for sampling ratio $\frac{M}{N} = 0.1$ .

\textbf{Initial Reconstruction Sub-network}. Given the compressed measurements, traditional BCS methods use the MMSE linear estimation to obtain the initial reconstructed signal
\begin{eqnarray} \label{eq4}
{\tilde x_j} = {\tilde \Phi _B}{y_j}
\end{eqnarray}
\begin{eqnarray}\label{eq5}
{\tilde \Phi _B} = {R_{xx}}\Phi _B^T{\left( {{\Phi _B}{R_{xx}}\Phi _B^T} \right)^{ - 1}}
\end{eqnarray}
where ${R_{xx}}$ is the autocorrelation function of the input signal. Obviously, ${\tilde \Phi _B}$ is a ${B^2} \times {n_B}$ matrix. Similar to the sampling process, we also use a convolution layer to mimic the initial reconstruction process. Compared with previous BCS method~\cite{rf1}, the matrix ${\tilde \Phi _B}$ is learned automatically in our network instead of computing by the complicated MMSE linear estimation. The convolution outputs of an image block in the sampling layer is a ${n_B} \times 1$  vector, so the size of the convolution filter in the initial reconstruction layer is $1 \times 1 \times {n_B}$. We use $1 \times 1$ stride convolution to reconstruct each block. Since this layer is used to mimic Eq. (\ref{eq4}), the biases is also ignored. In summary, we use ${B^2}$ convolution filters of size $1 \times 1 \times {n_B}$ to obtain each reconstructed block. However, the reconstructed output of each block is still a vector. To get the initial reconstructed image, we design a combination layer, which contains a reshape operator and a concatenation operator. This layer first reshapes each ${B^2} \times 1$ reconstructed vector to a $B \times B$ block, then concatenate the blocks to get the reconstructed image.

\textbf{Deep Reconstruction Sub-network}. As show in the upper of Fig. 2, there is a non-linear reconstruction process after getting the initial solution. In this paper, we design a deep sub-network, called as deep reconstruction sub-network, to achieve this function. It contains $m$ layers where the layers except the first and the last are of the same type: \emph{d} filters of the size $f \times f \times d$ , where a filter operates on a $f \times f$ spatial region across $d$ channels (feature maps). The first layer of the deep reconstruction sub-network operates on the initial reconstructed output, so that it has \emph{d} filters of size $f \times f \times 1$ . The last layer, which outputs the final image estimation, consists of a single filter of size $f \times f \times d$ . In our experiment, we set $d = 64$ and $f = 3$.

Finally, these three sub-networks, i.e. compressed sampling, initial reconstruction and deep reconstruction, form a compressed sensing based end-to-end deep networks. We name the proposed method as CSNet.

\subsection{Training}

Given the input image \emph{x}, our goal is to obtain the highly compressed measurement \emph{y} with the compressed sampling sub-network, and then accurately recover it to the original input image \emph{x} with the reconstruction sub-network. Since the sampling sub-network and the reconstruction sub-network form an end-to-end network \emph{f}, they can be trained together and do not need to be concerned with what the compressed measurement \emph{y} is. Therefore, the input and the label are all image \emph{x} itself for training our CSNet. Then the training dataset can be represented as $\left\{ {{x_i},{x_i}} \right\}_i^N$ . Following most of deep learning based image restoration methods, the mean square error is adopted as the cost function of our network. The optimization objective is represented as
\begin{eqnarray}
\min \frac{1}{{2N}}\sum\nolimits_{i = 1}^N {\left\| {f\left( {{x_i};\theta } \right) - {x_i}} \right\|} _F^2
\end{eqnarray}
where $\theta $ are the network parameters needed to be trained, $f\left( {{x_i};\theta } \right)$  is the final CS reconstructed output with respect to image ${x_i}$ . It should be noted that we train the compressed sampling sub-network and the reconstruction sub-network together, but they can be used independently. Furthermore, we only use the Rectified Linear Unit (ReLU) as activation function after each convolution layer in the deep reconstruction sub-network. Adaptive moment estimation (Adam)~\cite{rf21} is used to optimize all network parameters.

\begin{figure*}[t]
\centering
%\subfloat[sampling rate = 0.1]{%
\includegraphics[width=1\textwidth]{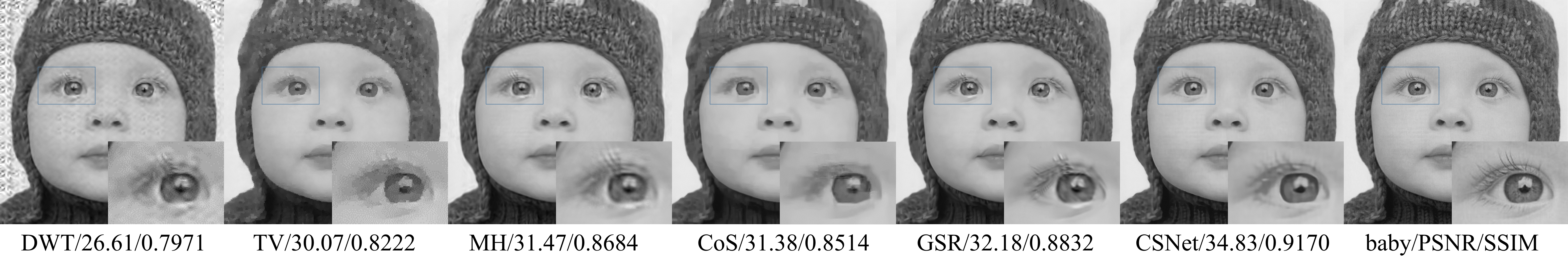}%}\vfill
%\subfloat[sampling rate = 0.3]{%
%\includegraphics[width=1\textwidth]{butterfly.png}}
%\begin{minipage}[b]{1\linewidth}
%  \centerline{\epsfig{figure=baby.eps,width=20.58cm}}
%  % \vspace{1.5cm}
%\end{minipage}
\vspace{-15pt}
\caption{\small {Visual quality comparison of image CS recovery on image baby from Set5~\cite{rf23} in the case of sampling ratio = 0.1.}}
\vspace{-5pt}
\label{fig:fig3}
\end{figure*}

%\begin{comment}
\begin{figure*}[t]
\setlength{\abovecaptionskip}{-10pt}
\setlength{\belowcaptionskip}{0pt}
\centering
%\subfloat[]{%
\includegraphics[width=1\textwidth]{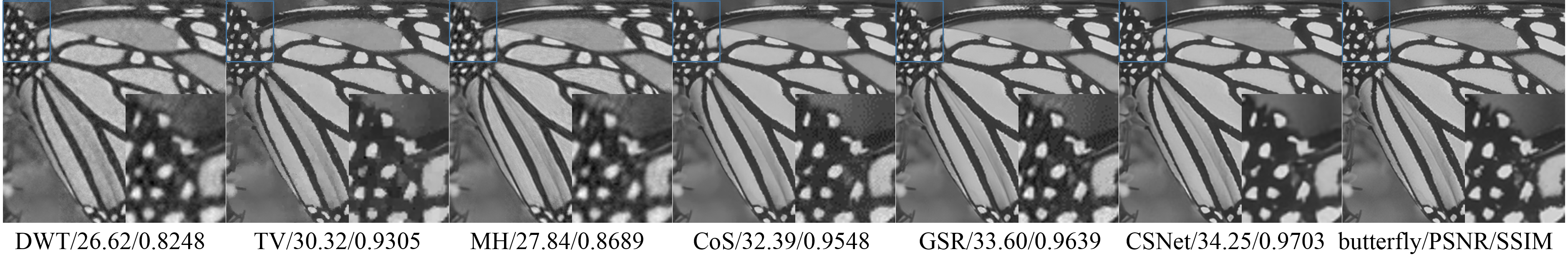}
%\begin{minipage}[b]{1\linewidth}
%  \centerline{\epsfig{figure=baby.eps,width=20.58cm}}
%  % \vspace{1.5cm}
%\end{minipage}
\vspace{-25pt}
\caption{\small {Visual quality comparison of image CS recovery on image butterfly from Set5~\cite{rf23} in the case of sampling ratio = 0.3.}}
\label{fig:fig4}
\vspace{-10pt}
\end{figure*}
%\end{comment}

\begin{table*}[t]\footnotesize
\centering
%\begin{center}
\caption{\small {PSNR$\backslash$SSIM$\backslash$running time comparisons with various algorithms on Set5~\cite{rf23}}}\label{tab:tab1}
%\vspace{-5pt}
\begin{tabular}{|l|l|l|l|l|l|l|}
\hline 
Alg. & baby & bird & butterfly & head & woman & Avg.\\ 
\hline 
\multicolumn{7}{|c|}{Sampling Ratio (M/N) 0.1 (PSNR$\backslash$SSIM$\backslash$running time)}\\ 
\hline 
DWT & 26.61$\backslash$0.7971$\backslash$33.87 & 29.82$\backslash$0.8692$\backslash$11.48 & 22.02$\backslash$0.7072$\backslash$9.93 & 19.82$\backslash$0.6501$\backslash$8.21 & 25.41$\backslash$0.8163$\backslash$10.40 & 24.74$\backslash$0.7680$\backslash$14.78\\ 
\hline 
TV & 30.07$\backslash$0.8222$\backslash$19.00 & 28.45$\backslash$0.8192$\backslash$4.07 & 21.65$\backslash$0.7481$\backslash$3.38 & 30.02$\backslash$0.7350$\backslash$4.42 & 25.15$\backslash$0.8079$\backslash$3.61 & 27.07$\backslash$0.7865$\backslash$6.90\\ 
\hline 
MH & 31.47$\backslash$0.8684$\backslash$54.58 & 31.74$\backslash$0.8947$\backslash$20.62 & 22.55$\backslash$0.7412$\backslash$14.29 & 30.64$\backslash$0.7546$\backslash$13.61 & 26.43$\backslash$0.8467$\backslash$15.43 & 28.57$\backslash$0.8211$\backslash$23.70\\ 
\hline 
CoS & 31.38$\backslash$0.8514$\backslash$\tiny 14044.44 & 31.92$\backslash$0.8973$\backslash$ \tiny 4688.28 & 25.40$\backslash$0.8737$\backslash$\tiny 5095.94 & 31.24$\backslash$0.7604$\backslash$\tiny 3548.57 & 27.82$\backslash$0.8783$\backslash$\tiny 3640.34 & 29.55$\backslash$0.8522$\backslash$\tiny 6203.51\\ 
\hline 
GSR & 32.18$\backslash$0.8832$\backslash$852.50 & 34.77$\backslash$0.9411$\backslash$276.54 & 23.78$\backslash$0.8279$\backslash$215.12 & 31.33$\backslash$0.7717$\backslash$215.85 & 27.88$\backslash$0.9029$\backslash$244.18 & 29.99$\backslash$0.8654$\backslash$360.84\\ 
\hline 
CSNet & \textbf{34.83}$\backslash$\textbf{0.9170}$\backslash$\textbf{0.05} & \textbf{35.15}$\backslash$\textbf{0.9476}$\backslash$\textbf{0.05} & \textbf{28.01}$\backslash$\textbf{0.9018}$\backslash$\textbf{0.04} & \textbf{33.26}$\backslash$\textbf{0.8208}$\backslash$\textbf{0.04} & \textbf{30.23}$\backslash$\textbf{0.9203}$\backslash$\textbf{0.02} & \textbf{32.30}$\backslash$\textbf{0.9015}$\backslash$\textbf{0.04}\\ 
\hline 
%Sampling Rate (M/N) 0.2 (PSNR$\backslash$SSIM) & • & • & • & • & • & • \\ 
\multicolumn{7}{|c|}{Sampling Ratio (M/N) 0.2 (PSNR$\backslash$SSIM$\backslash$running time)}\\
\hline 
DWT & 34.67$\backslash$0.9207$\backslash$24.36 & 36.15$\backslash$0.9489$\backslash$8.16 & 24.99$\backslash$0.7930$\backslash$6.74 & 28.00$\backslash$0.7955$\backslash$4.56 & 30.35$\backslash$0.9163$\backslash$8.72 & 30.83$\backslash$0.8749$\backslash$10.51\\ 
\hline 
TV & 32.60$\backslash$0.8821$\backslash$10.76 & 32.39$\backslash$0.9027$\backslash$3.37 & 26.86$\backslash$0.8848$\backslash$2.84 & 31.99$\backslash$0.7963$\backslash$2.42 & 28.40$\backslash$0.8884$\backslash$2.46 & 30.45$\backslash$0.8709$\backslash$4.37\\ 
\hline 
MH & 34.80$\backslash$0.9223$\backslash$39.89 & 36.35$\backslash$0.9503$\backslash$13.74 & 25.81$\backslash$0.8324$\backslash$21.60 & 32.98$\backslash$0.8224$\backslash$17.67 & 30.47$\backslash$0.9130$\backslash$17.92 & 32.08$\backslash$0.8881$\backslash$22.17\\ 
\hline 
CoS & 34.15$\backslash$0.9074$\backslash$\tiny 8025.87 & 36.77$\backslash$0.9554$\backslash$\tiny 2698.52 & 29.66$\backslash$0.9323$\backslash$\tiny 3469.95 & 31.91$\backslash$0.8043$\backslash$\tiny 5035.38 & 31.56$\backslash$0.9342$\backslash$\tiny 2550.58 & 32.81$\backslash$0.9067$\backslash$\tiny 4356.06\\ 
\hline 
GSR & 35.35$\backslash$0.9315$\backslash$828.85 & \textbf{40.03}$\backslash$0.9764$\backslash$264.59 & 29.29$\backslash$0.9337$\backslash$221.37 & 33.26$\backslash$0.8314$\backslash$218.75 & 32.92$\backslash$0.9555$\backslash$236.01 & 34.17$\backslash$0.9257$\backslash$353.91\\ 
\hline 
CSNet & \textbf{37.70}$\backslash$\textbf{0.9563}$\backslash$\textbf{0.06} & 39.78$\backslash$\textbf{0.9804}$\backslash$\textbf{0.02} & \textbf{31.79}$\backslash$\textbf{0.9523}$\backslash$\textbf{0.02} & \textbf{35.08}$\backslash$\textbf{0.8763}$\backslash$\textbf{0.01} & \textbf{33.82}$\backslash$\textbf{0.9603}$\backslash$\textbf{0.02} & \textbf{35.63}$\backslash$\textbf{0.9451}$\backslash$\textbf{0.02}\\ 
\hline 
%Sampling Rate (M/N) 0.3 (PSNR$\backslash$SSIM) & • & • & • & • & • & • \\
\multicolumn{7}{|c|}{Sampling Ratio (M/N) 0.3 (PSNR$\backslash$SSIM$\backslash$running time)}\\
\hline 
DWT & 36.50$\backslash$0.9430$\backslash$16.33 & 38.77$\backslash$0.9677$\backslash$5.65 & 26.62$\backslash$0.8247$\backslash$4.05 & 33.44$\backslash$0.8494$\backslash$2.43 & 32.74$\backslash$0.9400$\backslash$7.46 & 33.61$\backslash$0.9050$\backslash$7.18\\ 
\hline 
TV & 34.45$\backslash$0.9170$\backslash$8.26 & 35.06$\backslash$0.9392$\backslash$2.82 & 30.32$\backslash$0.9305$\backslash$2.11 & 33.27$\backslash$0.8401$\backslash$2.00 & 30.67$\backslash$0.9266$\backslash$2.09 & 32.75$\backslash$0.9107$\backslash$3.46\\ 
\hline 
MH & 36.50$\backslash$0.9430$\backslash$29.93 & 38.77$\backslash$0.9677$\backslash$10.12 & 27.84$\backslash$0.8689$\backslash$14.87 & 34.28$\backslash$0.8596$\backslash$10.73 & 32.90$\backslash$0.9397$\backslash$14.03 & 34.06$\backslash$0.9158$\backslash$15.93\\ 
\hline 
CoS & 35.88$\backslash$0.9326$\backslash$\tiny 6635.08 & 39.60$\backslash$0.9733$\backslash$\tiny 2230.49 & 32.39$\backslash$0.9548$\backslash$\tiny 2855.96 & 33.78$\backslash$0.8496$\backslash$\tiny 5245.95 & 33.68$\backslash$0.9553$\backslash$\tiny 1650.00 & 35.07$\backslash$0.9331$\backslash$\tiny 3723.50\\ 
\hline 
GSR & 37.32$\backslash$0.9538$\backslash$845.79 & 42.84$\backslash$0.9861$\backslash$272.09 & 33.60$\backslash$0.9639$\backslash$232.55 & 34.64$\backslash$0.8705$\backslash$248.28 & 35.73$\backslash$0.9720$\backslash$271.82 & 36.83$\backslash$0.9492$\backslash$374.10\\ 
\hline 
CSNet & \textbf{39.69}$\backslash$\textbf{0.9731}$\backslash$\textbf{0.05} & \textbf{42.97}$\backslash$\textbf{0.9900}$\backslash$\textbf{0.02} & \textbf{34.25}$\backslash$\textbf{0.9703}$\backslash$\textbf{0.02} & \textbf{36.35}$\backslash$\textbf{0.9061}$\backslash$\textbf{0.01} & \textbf{36.25}$\backslash$\textbf{0.9753}$\backslash$\textbf{0.02} & \textbf{37.90}$\backslash$\textbf{0.9630}$\backslash$\textbf{0.02}\\ 
\hline 
%Sampling Rate (M/N) 0.4 (PSNR$\backslash$SSIM) & • & • & • & • & • & • \\
\multicolumn{7}{|c|}{Sampling Ratio (M/N) 0.4 (PSNR$\backslash$SSIM$\backslash$running time)}\\
\hline 
DWT & 37.97$\backslash$0.9575$\backslash$12.81 & 40.84$\backslash$0.9777$\backslash$3.76 & 28.17$\backslash$0.8528$\backslash$4.66 & 35.03$\backslash$0.8800$\backslash$4.61 & 34.60$\backslash$0.9565$\backslash$5.54 & 35.32$\backslash$0.9249$\backslash$6.27\\ 
\hline 
TV & 36.06$\backslash$0.9404$\backslash$6.76 & 37.56$\backslash$0.9615$\backslash$2.82 & 33.66$\backslash$0.9574$\backslash$1.80 & 34.45$\backslash$0.8724$\backslash$1.67 & 32.72$\backslash$0.9499$\backslash$1.67 & 34.89$\backslash$0.9363$\backslash$2.95\\ 
\hline 
MH & 37.97$\backslash$0.9575$\backslash$29.71 & 40.84$\backslash$0.9777$\backslash$8.99 & 29.45$\backslash$0.8922$\backslash$30.80 & 35.31$\backslash$0.8866$\backslash$19.32 & 34.70$\backslash$0.9544$\backslash$13.52 & 35.65$\backslash$0.9337$\backslash$20.47\\ 
\hline 
CoS & 37.59$\backslash$0.9524$\backslash$\tiny 5177.50 & 41.96$\backslash$0.9832$\backslash$\tiny 1566.75 & 34.81$\backslash$0.9685$\backslash$\tiny 1850.93 & 35.14$\backslash$0.8839$\backslash$\tiny 4308.08 & 35.79$\backslash$0.9693$\backslash$\tiny 1468.03 & 37.06$\backslash$0.9515$\backslash$\tiny 2874.26\\ 
\hline 
GSR & 39.08$\backslash$0.9682$\backslash$981.18 & 45.12$\backslash$0.9910$\backslash$460.48 & 36.23$\backslash$0.9754$\backslash$462.25 & 35.72$\backslash$0.8974$\backslash$271.72 & 37.91$\backslash$0.9810$\backslash$287.80 & 38.81$\backslash$0.9626$\backslash$492.68\\ 
\hline 
CSNet & \textbf{41.58}$\backslash$\textbf{0.9830}$\backslash$\textbf{0.09} & \textbf{45.52}$\backslash$\textbf{0.9942}$\backslash$\textbf{0.03} & \textbf{36.48}$\backslash$\textbf{0.9801}$\backslash$\textbf{0.02} & \textbf{37.45}$\backslash$\textbf{0.9264}$\backslash$\textbf{0.02} & \textbf{38.44}$\backslash$\textbf{0.9847}$\backslash$\textbf{0.03} & \textbf{39.89}$\backslash$\textbf{0.9736}$\backslash$\textbf{0.04}\\ 
\hline 
%Sampling Rate (M/N) 0.5 (PSNR$\backslash$SSIM) & • & • & • & • & • & • \\
\multicolumn{7}{|c|}{Sampling Ratio (M/N) 0.5 (PSNR$\backslash$SSIM$\backslash$running time)}\\
\hline 
DWT & 39.40$\backslash$0.9683$\backslash$10.38 & 42.60$\backslash$0.9839$\backslash$3.11 & 29.76$\backslash$0.8794$\backslash$2.48 & 36.07$\backslash$0.9048$\backslash$2.21 & 36.54$\backslash$0.9683$\backslash$4.87 & 36.87$\backslash$0.9409$\backslash$4.61\\ 
\hline 
TV & 37.55$\backslash$0.9568$\backslash$5.90 & 39.66$\backslash$0.9742$\backslash$1.76 & 36.20$\backslash$0.9714$\backslash$2.31 & 35.57$\backslash$0.9007$\backslash$1.89 & 34.78$\backslash$0.9668$\backslash$1.67 & 36.75$\backslash$0.9540$\backslash$2.71\\ 
\hline 
MH & 39.40$\backslash$0.9683$\backslash$29.45 & 42.60$\backslash$0.9839$\backslash$9.21 & 31.16$\backslash$0.9131$\backslash$13.35 & 36.35$\backslash$0.9104$\backslash$10.01 & 36.54$\backslash$0.9652$\backslash$13.38 & 37.21$\backslash$0.9482$\backslash$15.08\\ 
\hline 
CoS & 39.10$\backslash$0.9651$\backslash$\tiny 4906.03 & 43.86$\backslash$0.9884$\backslash$\tiny 2437.14 & 37.27$\backslash$0.9781$\backslash$\tiny 3103.51 & 36.23$\backslash$0.9060$\backslash$\tiny 7803.89 & 37.58$\backslash$0.9784$\backslash$\tiny 2389.08 & 38.81$\backslash$0.9632$\backslash$\tiny 4127.93\\ 
\hline 
GSR & 40.83$\backslash$0.9781$\backslash$\tiny 1010.73 & \textbf{47.06}$\backslash$0.9938$\backslash$322.31 & \textbf{38.62}$\backslash$\textbf{0.9826}$\backslash$259.95 & 36.84$\backslash$0.9207$\backslash$261.22 & \textbf{39.91}$\backslash$0.9870$\backslash$286.17 & 40.65$\backslash$0.9724$\backslash$428.08\\ 
\hline 
CSNet & \textbf{43.00}$\backslash$\textbf{0.9879}$\backslash$\textbf{0.06} & 46.26$\backslash$\textbf{0.9949}$\backslash$\textbf{0.02} & 37.53$\backslash$0.9821$\backslash$\textbf{0.01} & \textbf{38.28}$\backslash$\textbf{0.9386}$\backslash$\textbf{0.02} & 39.74$\backslash$\textbf{0.9885}$\backslash$\textbf{0.02} & \textbf{40.96}$\backslash$\textbf{0.9784}$\backslash$\textbf{0.03}\\\hline 
\end{tabular} 
%\end{center}
\vspace{-15pt}
\end{table*}

\section{Experimental Results}

In this section, we evaluate the performance of the proposed CSNet for CS reconstruction. We first describe the datasets used for training and testing. Next, some training details are given. Finally, we show the quantitative and qualitative comparisons with five state-of-the-art methods.

\subsection{Datasets for Training and Testing}

We use the training set (200 images) and test set (200 images) of the BSDS500 database~\cite{rf22} for training, and its validation set (100 images) for validation. We set the batch size as $96 \times 96$, and use data augmentation (rotation or flip) to prepare training data. To reduce memory usage, we only keep $64 \times 1400$ patches, which generate good enough performance for comparison. For benchmark, we use two test datasets: Set5~\cite{rf23} (5 images) and Set14~\cite{rf24} (14 images) that are widely used for benchmark in other works. Note that the test images are strictly separate from the training datasets.

\subsection{Training Details}

The basic network parameters have been described in section 3.1. We use the method described in~\cite{rf25} to initialize weights, which is a theoretically sound procedure for networks utilizing rectified linear units. For other hyper-parameters of Adam, we set the exponential decay rates for the first and second moment estimate to 0.9 and 0.999, respectively. We train our model for 100 epochs and each epoch iterates 1400 times with batch size 64. The learning rate of the first 50 epochs is 0.001, the 51 to 80 epochs is 0.0001, while that of the other 20 epochs is 0.00001. We found that if we meticulously choose the training output, a better result will be obtained. However, for the sake of simplicity, we just report the test results by the hundredth training epoch, which achieves a good enough performance for comparison. We implement our model using the MatConvNet package~\cite{rf26}. Training takes roughly five hours on a GPU Titan X.

\subsection{Comparisons with State-of-the-Art Methods}

Our proposed algorithm is compared with five representative CS recovery methods in the literature, i.e., wavelet method (DWT)~\cite{rf27}, total variation (TV) method~\cite{rf7}, multi-hypothesis(MH) method~\cite{rf10}, collaborative sparsity (CoS) method~\cite{rf28} and group sparse representation (GSR) method~\cite{rf11}. All these methods are BCS methods, and the block size is also 32. The implementation codes are downloaded from the authors’ websites and the default parameter settings are used in our experiments. To evaluate the performance of each algorithm, we investigate five different sampling ratio from 0.1 to 0.5 with assessment criteria PSNR, SSIM and running time. All the test experiments are implemented in Matlab 2015a on Windows 7 system, and runs on desktop computer with 4 cores CPU at 3.4 GHz and 12 GB RAM. Both quantitative and qualitative comparisons are given. The comparisons with various algorithms on Set5 in case of 0.1-0.5 measurements are provided in Table 1. Our proposed CSNet achieves the highest PSNR and SSIM and the least running time among all comparative algorithms. Compare to GSR, our CSNet can improve roughly 2.31 dB, 1.46 dB, 1.07 dB, 1.08 dB and 0.31 dB on average with respect to 0.1-0.5 sampling ratio, respectively. Table 2 shows the average PSNR, SSIM and running time of various algorithms on Set14. On this test dataset, our CSNet can improve roughly 3.91 dB, 3.71 dB, 2.60 dB, 1.72 dB and 0.37 dB on average, in comparison with DWT, TV, MH, CoS and GSR, respectively. Both the PSNR and SSIM values of Table 1 and Table 2 demonstrate our proposed CSNet obtains the best performance. Furthermore, our method run fastest as show in Fig.1, Table 1 and Table 2, which is very important for real time applications. To get better reconstruction results, we can increase the depth of the deep reconstruction sub-network, which is called “deeper is better” in the literature, or the number of neurons of each layer. In our experiments, we have already implemented a deeper CSNet with 10 layers of deep reconstruction sub-network, and a fatter CSNet with $d = 128$. They all show PSNR and SSIM gain. Some visual results of the recovered images by various algorithms are presented in Fig.3 and Fig.4. Our proposed CSNet preserves much sharper edges and finer details, showing much clearer and better visual results than other competing methods.

\begin{table*}[t]\footnotesize
%\setlength{\abovecaptionskip}{0pt}
%\setlength{\belowcaptionskip}{-16pt}
%\setlength{\belowcaptionskip}{-30pt}
%\begin{center}
\vspace{-3pt}
\centering
\caption{\small {Average PSNR$\backslash$SSIM$\backslash$running time comparisons with various algorithms on Set14~\cite{rf24}}}\label{tab:tab2}
\begin{tabular}{|l|l|l|l|l|l|l|}
\hline 
Alg. & Sampling ratio 0.1 & Sampling ratio 0.2 & Sampling ratio 0.3 & Sampling ratio 0.4 & Sampling ratio 0.5 & Avg.\\ 
\hline 
DWT & 24.16$\backslash$0.6798$\backslash$32.46 & 28.13$\backslash$0.7882$\backslash$24.43 & 30.38$\backslash$0.8389$\backslash$15.75 & 31.99$\backslash$0.8753$\backslash$11.71 & 33.54$\backslash$0.9044$\backslash$9.37 & 29.64$\backslash$0.8173$\backslash$18.70\\ 
\hline 
TV & 25.24$\backslash$0.6887$\backslash$16.02 & 28.07$\backslash$0.7844$\backslash$13.34 & 30.12$\backslash$0.8424$\backslash$10.25 & 32.03$\backslash$0.8837$\backslash$8.42 & 33.84$\backslash$0.9148$\backslash$7.67 & 29.84$\backslash$0.8228$\backslash$11.14\\ 
\hline 
MH & 26.38$\backslash$0.7282$\backslash$64.22 & 29.47$\backslash$0.8237$\backslash$60.33 & 31.37$\backslash$0.8694$\backslash$52.35 & 33.03$\backslash$0.9009$\backslash$44.18 & 34.52$\backslash$0.9239$\backslash$42.11 & 30.95$\backslash$0.8492$\backslash$52.64\\ 
\hline 
CoS & 27.20$\backslash$0.7433$\backslash$\tiny 18698.82 & 30.07$\backslash$0.8278$\backslash$\tiny 17762.84 & 32.03$\backslash$0.8732$\backslash$\tiny 17314.04 & 34.00$\backslash$0.9084$\backslash$\tiny 15371.97 & 35.84$\backslash$0.9314$\backslash$\tiny 14956.68 & 31.83$\backslash$0.8568$\backslash$\tiny 16820.87\\ 
\hline 
GSR & 27.50$\backslash$0.7705$\backslash$883.72 & 31.22$\backslash$0.8642$\backslash$883.08 & 33.74$\backslash$0.9071$\backslash$815.99 & 35.78$\backslash$0.9336$\backslash$799.38 & \textbf{37.66}$\backslash$0.9522$\backslash$815.61 & 33.18$\backslash$0.8855$\backslash$839.56\\ 
\hline 
CSNet & \textbf{28.91}$\backslash$\textbf{0.8119}$\backslash$\textbf{0.12} & \textbf{31.86}$\backslash$\textbf{0.8908}$\backslash$\textbf{0.12} & \textbf{34.00}$\backslash$\textbf{0.9276}$\backslash$\textbf{0.12} & \textbf{35.95}$\backslash$\textbf{0.9495}$\backslash$\textbf{0.15} & 37.05$\backslash$\textbf{0.9607}$\backslash$\textbf{0.14} & \textbf{33.55}$\backslash$\textbf{0.9081}$\backslash$\textbf{0.13}\\\hline 
\end{tabular}
%\end{center}
%\setlength{\textfloatsep}{-10pt}
\vspace{-15pt}
\end{table*} 
\section{Conclusion}

In this paper, we use deep learning to solve the two most important CS issues, i.e. designing a sampling operator and developing a fast nonlinear reconstruction algorithm. We design a deep network that consists of three sub-networks: compressed sampling, initial reconstruction and deep reconstruction, which has high relationship with traditional block compressed sensing smooth projected Landweber algorithm. By designing a sampling sub-network, the sampling operator can be learned automatically, which avoids complicated artificial designs. Given the sampling measurement, the reconstruction sub-network can efficiently recover the original image. Experimental results show that the proposed CSNet achieves significant performance improvements over several current state-of-the-art methods, and runs in real time. In future work, we will take residual learning into account to further improve reconstruction performance and running time.

\bibliographystyle{IEEEbib}
\scriptsize
\setlength{\bibsep}{0.9ex}
\bibliography{camera-ready_icme2017template}

\end{document}